\documentclass[runningheads]{llncs}
\pdfoutput=1
\usepackage{graphicx}
\usepackage{amsfonts} 
\usepackage{amsmath} 
\usepackage{multicol}
\usepackage{lmodern}
\usepackage{microtype}
\usepackage{subcaption}
\usepackage{algorithm}
\usepackage{algpseudocode}
\algrenewcommand\algorithmicindent{1.0em}%

\usepackage{mathtools}

\let\given\givenbase

\DeclarePairedDelimiterX{\infdivx}[2]{(}{)}{%
  #1\;\delimsize\|\;#2%
}
\newcommand{\kl}{\operatorname{KL}\infdivx}
\newcommand{\dist}{\operatorname{d}\infdivx}

\usepackage{tikz}
\usetikzlibrary{
  calc,
  positioning,
  arrows.meta,
  fit,
  external,
}

\tikzset{
  edge/.style={
    ->,
    >=Latex,
    shorten >=5pt,
    shorten <=5pt,
    rounded corners,
  }
}

\usepackage{pgfplots}
\usepackage{pgfplotstable}
\usepgfplotslibrary{groupplots}
\usepgfplotslibrary{
  colorbrewer,
  groupplots,
}

\pgfplotsset{%
  compat=1.17,
  grid style=dashed,
  ymajorgrids=true,
  cycle list/Dark2,
  dark marker list/.style={cycle multiindex* list={
    Dark2\nextlist
    mark list\nextlist
    mark options={fill=.!75}
  }},
  dark list/.style={cycle multiindex* list={
    Dark2\nextlist
    fill=.!75
  }},
}

\usepackage{cite}

\begin{document}
\title{RepFair-GAN: Mitigating Representation Bias in GANs Using Gradient Clipping%
}
\titlerunning{RepFair-GAN: Mitigating Representation Bias in GANs}
%
\author{
 Patrik Joslin Kenfack\inst{1} \and
 Kamil Sabbagh\inst{1} \and
 Adín Ramírez Rivera\inst{2}\and
 Adil Khan\inst{1}
}
\authorrunning{Kenfack et al.}
%
\institute{
Machine Learning and Knowledge Representation Lab,\\ Innopolis University, Innopolis, Russia\\ 
\email{\{p.kenfack,k.sabbagh\}@innopolis.university, a.khan@innopolis.ru}\\ \and
Departments of Informatics, University of Oslo, Oslo, Norway\\ 
\email{adinr@uio.no}\\ 
}
\maketitle     
%
%
%
%
%

\begin{abstract}
Fairness has become an essential problem in many domains of Machine Learning (ML), such as classification, natural language processing, and Generative Adversarial Networks (GANs). In this research effort, we study the unfairness of GANs. We formally define a new fairness notion for generative models in terms of the distribution of generated samples sharing the same protected attributes (gender, race, etc.). The defined fairness notion (\textit{representational fairness}) requires the distribution of the sensitive attributes at the test time to be uniform, and, in particular for GAN model, we show that this fairness notion is violated even when the dataset contains equally represented groups, i.e., the generator favors generating one group of samples over the others at the test time.    
In this work, we shed light on the source of this representation bias in GANs along with a straightforward method to overcome this problem. We first show on two widely used datasets (MNIST, SVHN) that when the norm of the gradient of one group is more important than the other during the discriminator's training, the generator favors sampling data from one group more than the other at test time. We then show that controlling the groups' gradient norm by performing \textit{group-wise} gradient norm clipping in the discriminator during the training leads to a more fair data generation in terms of \textit{representational fairness} compared to existing models while preserving the quality of generated samples. 

\end{abstract}
 
\section{Introduction}
The increasing use of Machine Learning (ML) models to make high-stake decisions in areas like hiring, college admission and criminal justice has raised concerns about the fairness of decisions made by these models~\cite{barocas2016big}. In fact, the data collected to train these models may also incorporate biases or historical discrimination from the real world. As ML models are good at discovering patterns in the data, they can also learn and perpetuate these biases, leading to discriminatory outcomes against subgroups in the population (minorities)~\cite{mehrabi2021survey,caton2020fairness,kenfack2021impact}. Several efforts have been made to  define and quantify fairness mathematically~\cite{dwork2012fairness,hardt2016equality,corbett2017algorithmic,berk2012criminal}. The proposed definitions of fairness aim to equalize models' performances or error rates across groups present in the dataset. Algorithms to achieve these fairness notions can be classified into three main categories: pre-processing~\cite{zemel2013learning,kamiran2012data,madras2018learning,kenfack2021adversarial}, in-processing~\cite{zhang2018mitigating}, and post-processing~\cite{hardt2016equality}, depending on whether biases are mitigated at the data level, during model training or after model training, respectively. Biases can originate from different sources such as the real world, user interactions, and algorithmic bias~\cite{mehrabi2021survey}. The most common type of bias in the real world is the \textit{representation bias}, which occurs when certain groups or subgroups are under-represented in the dataset. The model will tend to be biased in favor of the well-represented groups. One way to mitigate representation bias is to collect more data from under-represented groups, but these data are difficult to collect or are not available in most cases. To overcome this limitation, one can rely on generative models to augment the dataset with synthetic data~\cite{tanaka2019data,mariani2018bagan}.  

Generative Adversarial Network (GAN) is considered as the most prominent technique for learning the underlying distribution of data~\cite{goodfellow2014generative}. The model consists of two competing networks: the generator $G$, which captures the distribution by sampling as realistic data as possible from the noize vector, and the discriminator $D$, which distinguishes between samples from $G$ (fake sample) and the data. The goal of $G$ is to defeat the discriminator to distinguish between fake and real samples. Once this goal is achieved, $G$ can be used to generate synthetic data. GAN models and their variants have shown incredible results in synthetic image generation~\cite{karras2017progressive,karras2019style} and music generation~\cite{dong2018musegan}. As GAN models are increasingly used as data augmentation tools, it is crucial to ensure that the sampling process is fair and free from any type of bias. Previous works have shown that GANs can also generate biased data~\cite{kenfack2021fairness,xu2018fairgan,tan2020improving} by encoding disparities in the data and generating imbalance data, i.e., synthetic data close to realistic data but are less representative of the underlying population. In essence, 
work done in~\cite{kenfack2021fairness, tan2020improving} has shown that when certain groups are under-represented (minorities) in the data, the generator tends to favor well-represented groups at testing time, leading to skewed data that can result in unfairness in downstream tasks~\cite{mehrabi2021survey}. 

This paper addresses the issue of representation bias in GAN; we provide insights into why the generator favors one group at the test time. We observe through experiments that when there are disparities in the magnitude of the gradients of different groups in discriminator during training, the generator tends to favor the groups with larger magnitude at test time, i.e., the generate samples points from one group more often than the other. Based on the previous observation, we propose a training process that allows the generator to uniformly generate data from different groups at the testing time by controlling the gradient of each group during the training. We achieve this by clipping the norm of the gradient of groups during the training of the discriminator; This avoids the discriminator to be more powerful in certain groups than the other, and as such, it enforces the generator to not mainly converges towards the group with which it can easily defeat the discriminator. Our method can be incorporated easily in many GAN variants as the discriminator's gradient is readily available in general.     

The contributions of this paper are as follows:
\begin{itemize}
    \item We formalize a fairness notion for generative models (\textit{representational fairness}) that promotes the uniformity of the sensitive attributes in the samples generated at the test time. In the context of GAN, this notion is achieved if, at the test time, the generator uniformly samples points from different groups present in the training dataset. In other words, a well-trained generator violates this notion if it is more likely to sample points from one group than the others.
    \item We conducted empirical studies to reveal the reason why GANs generator may favor one group over the others at the test time. We showed that the magnitude of the gradient of groups on the discriminator side during the training plays an important role in guiding the generator to sample groups uniformly at the test time. We considered the cases where group representations in the training dataset are balanced and imbalanced. Our results also showed that even when we train GAN on balanced datasets, the generator can still favor one group over the others at testing time.  
    \item We proposed Representational Fair-GAN (RepFair-GAN) an adaptation of GAN that clips the gradient norm group-wise to avoid one group dominating the training and to provide a more uniform (fair) data generation at the testing time.
    \item We demonstrated through intensive experiments that our method--yet simple-- is effective and can generate a more balanced dataset than the existing GANs models, even when the dataset is imbalanced.
\end{itemize}
The rest of this paper is organized as follows.  In Section~\ref{sec:related}, we discuss related work and the position of our work. In Sections~\ref{sec:method} and~\ref{sec:experiments},  we present empirical evidence of the effect of disparity in groups' gradient norms on the fairness of the generator and experimental results respectively. Section~\ref{ref:conclusion} is dedicated to the conclusion and future works.







\section{Related Work}
\label{sec:related}
The approaches closest in spirit to our work involve fair data generation using GANs. These approaches differ mainly in the fairness notion they aim to achieve or the type of bias they aim to mitigate. FairGAN~\cite{xu2018fairgan} uses one generator and two discriminators. As in the classic GAN, the goal of the first generator is to distinguish between fake and real samples while the goal of the second discriminator is to predict the sensitive attribute (ethnicity, gender) from fake and real samples. In a mini-max game optimization, the goal of the generator in FairGAN is to defeat the two discriminators in their respective objectives. Thus, FairGAN generates samples that are independent of the sensitive attribute to achieve \textit{statistical parity} in a downstream classifier.  Statistical parity~\cite{dwork2012fairness} is a fairness notion mainly used in classification tasks that promotes the independence between the classifier's positive outcome and the sensitive attribute, i.e., $P(Y=1 \given S=1) = P(Y=1 \given S=0)$ where $S$ is the random variable describing the sensitive attribute and $Y$ the classifier's output. 
Xu et al.~\cite{xu2019fairgan+} take a similar approach but add a classifier and consider three discriminators to improve the predictive performance of the generated samples while satisfying other notions of fairness such as equalized odds and equal opportunity~\cite{hardt2016equality}. In contrast to previous work, our work does not target any specific downstream task or fairness notion but focuses on fair sampling. As GANs are primarily used as a data augmentation tool, our goal is to mitigate the representation bias exhibited by the generator~\cite{kenfack2021fairness}, i.e., to improve the ability of GAN models to uniformly generate samples from different groups, even when these groups are not equally represented in the training data.   

More closely related to our work, Choi et al.~\cite{choi2020fair} use instance reweighing during the training of GANs such that over-represented data points are down-weighted while under-represented are up-weighted. Tan et al.~\cite{tan2020improving} propose a \textit{post-training} approach that considers already well-trained GANs, and shifts the distribution of the latent space in a way to provide a more fair sampling. In our work, we control the gradient flowing from the discriminator, in a way that it does not provide a bigger signal for one group than others. We clip the gradient norm group-wise, in a way that the magnitude of the gradient of one group will not enlarge during the training. Our empirical results demonstrate that our training process results in fairer vanilla GANs in terms of uniformity of the sampled data points, i.e., all groups are equally likely to be sampled by the generator at test time.  The advantages of our method are as follows: (1)~Gradient clipping is readily available and general, and as such, can be incorporated easily for many GAN variants. 
In our experiment, we use the vanilla GANs~\cite{goodfellow2014generative} and the conditional GAN (CGAN)~\cite{mirza2014conditional}. (2)~It preserves high-quality data generation with a carefully chosen clipping factor. (3)~When the groups are not equally represented in the training data, our training process leads to uniform data generation while preserving the data quality of both groups.

\section{RepFair-GAN}
\label{sec:method}
In this section, we provide more details on the problem statement with the evidence of biased data generation of the classic GANs model. We also demonstrate the influence of disparity in groups' gradient magnitude over the generation process and describe the training of our method (RepFair-GAN).

\subsection{Representational Fairness}
We consider a dataset $\mathcal{D}= \{\mathcal{X}, \mathcal{S}\}$ of size $N$ where $\mathcal{X}=\{x_i\}_{i=1}^N$ ; $x_i \in \mathbb{R}^d$ represents data points sampled from an unknown distribution $P_{\operatorname{data}}(\mathcal{X})$, and $\mathcal{S}=\{s_i\}_{i=1}^N$ the binary sensitive attribute associated with each point. Each data point $x_i$ can be an image and we assume the image contains sensitive attributes, like the gender of the face in the image.  The goal of a GANs is to estimate the underlying distribution $P_{\theta}(\mathcal{X})$ of the data, such that $P_{\theta}(\mathcal{X}) = P_{\operatorname{data}}(\mathcal{X})$. In the classic GAN, the optimization aims to minimize the Jensen-Shannon divergence between the two distribution. Given a well trained GAN on the observed data $\mathcal{D}$, its generator $g_{\theta}: \mathbb{R}^n \rightarrow \mathbb{R}^d$ can generate realistic data $D_{\theta} = \{g_{\theta}(\mathcal{Z})\}$    where $\mathcal{Z}=\{z_i\}_{i=1}^N$ ; $z_i \in \mathbb{R}^n$ represents latent space codes drawn from a given distribution $P(\mathcal{Z})$, generally assumed Gaussian. 

In addition, we consider a function $h: \mathcal{X} \rightarrow \mathcal{S}$ that predicts the sensitive attribute of the given image, e.g., predicting the gender of the face in an image or predicting the color of a digit on an image. We assume without loss of generality that the classifier $h$ is very accurate. Using $h$, we can analyze the distribution of sensitive attributes of the generated data points, i.e., $P(h(g_{\theta}(\mathcal{Z}))$. Our goal is to enforce fair sampling by making the distribution of the sensitive attributes over the generator $g_{\theta}$ close to uniform distribution, i.e., $\kl{P(h(g_{\theta}(\mathcal{Z}))}{\mathcal{U}(\mathcal{S})}=0$ where $\mathcal{U}$ is the uniform distribution, we refer to this fairness notion as \textit{representational fairness}. In essence, a biased generator has a higher probability of sampling data points from one group than the other. Formally, we define the "representational fairness" notion for generative models as follows:   
\begin{definition}
\label{def:representational-fairness}
A well-trained generator ($g_{\theta}$) is \textit{$\epsilon$-representationaly fair} with respect to the sensitive attributes $S$ if the distribution of $S$ over a set point randomly sampled from $g_{\theta}$,  $D_{\theta} = \{g_{\theta}(\mathcal{Z})\}$, is closed to the uniform distribution, i.e., for an $\epsilon \geq 0$,  $\; \dist{P(h(g_{\theta}(\mathcal{Z}))}{\mathcal{U}(\mathcal{S})} \leq \epsilon$. Where $\operatorname{d}$ is a distance over probability distributions, e.g., the KL-divergence;  $\mathcal{U}$ the uniform distribution; and $\epsilon$ the level of group balance achieved by $g_{\theta}$, i.e., the maximum distance from the uniform distribution obtained across multiples random sampling from $g_{\theta}$.  
\end{definition}
When $\epsilon = 0$, the generative model achieves uniformity over the sensitive attributes. The lower the value of $\epsilon$, the fairer the generator, while a high value of $\epsilon$ indicates an imbalanced representation of the sensitive attribute in $D_{\theta}$. 

\subsection{Biased data generation}
Previous works have studied biases in GANs~\cite{tan2020improving,kenfack2021fairness,choi2020fair}. These works have provided evidence that the distribution of generated data from different GAN models do not follow the uniform distribution over the sensitive attribute. They also provide evidence on how the biased data generated can lead to unfairness on downstream tasks such as classification, i.e., a classifier trained on synthetic data is more accurate on one group than another (minorities). The main reason for this behaviour stems from the fact that minority groups are underrepresented in the training data obtained by the GANs.     
However, most of these works considered settings where the training data contains imbalanced group distributions. In addition, we show in the experiment section that even when the training contains a balanced group representation, the generator still outputs non-uniform distribution at the test time. We observe through the experiments that there is a disparity in the magnitude of the gradient of each group during the training of the discriminator, and we hypothesize that this is the source of bias generation at the test time.

\subsection{The influence of the magnitude of the gradient vectors}
\subsubsection{Why disparity in groups' gradient norms causes unfair generation?} A question that naturally arises is why the magnitude of the gradient between groups is different during training and how this lead to a biased generator.

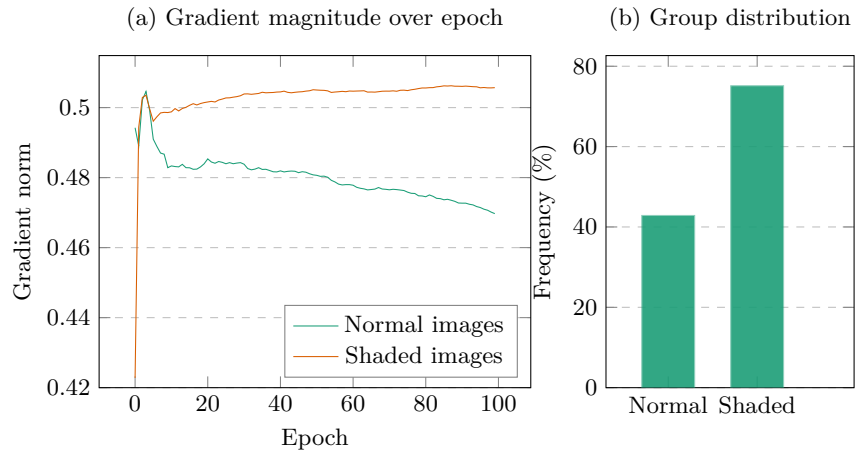
\begin{figure}[tb]
    \centering
    \begin{tikzpicture}
        \begin{groupplot}[
        group style={
                      group name=myplot,
                      group size= 2 by 1}, 
                      ]
        \nextgroupplot[
      legend style={ 
        draw=gray,  
        },  
    legend pos=south east,
    width=0.6\linewidth,
    height=6cm,
    xlabel={Epoch}, 
    ymin=.42,
    ylabel={Gradient norm},
    legend entries={Normal images,Shaded images},
    title={(a) Gradient magnitude over epoch}
    ]
    \addplot table [x=epoch, y=normal, col sep=comma]{images/csv/Vanilla-Gan_avg_magnitude.csv};
    \addplot table [x=epoch, y=shaded, col sep=comma]{images/csv/Vanilla-Gan_avg_magnitude.csv};
    
    \nextgroupplot[ 
        width=0.4\linewidth,
        height=6cm,
        xtick={0,1},
        xticklabels={Normal, Shaded},
        bar width = 20pt, 
        ymin=0,  
        xmax=2.09,
        xmin=-.7,
        ylabel={Frequency (\%)},
        title={(b) Group distribution}
        ]
            \addplot+ [ybar=.5cm, fill=Dark2-A, fill opacity=.9, draw opacity=0.5] table [x index=0, y=percentages, col sep=comma]{images/csv/Frequancy_of_each_image_type_normal_GAN.csv};
        \end{groupplot}
    \end{tikzpicture}
    \caption{Average gradient norms of different groups across epochs during the training of the vanilla GANs in the SVHN dataset. The difference in the magnitude of the gradients of each group enlarges during the training (a). After the training, the generator samples more data points from the group with a larger gradient during the training (b).}
    \label{fig:average_gradient_magni}
\end{figure} 
 In the mini-max training of GANs, the discriminator is simply a classifier that distinguishes samples provided by the generator, from real samples obtained from the training data. However, as described by Goodfellow et al.~\cite{goodfellow2014generative}, if the discriminator is too good, it may not provide enough information in the gradients (vanishing) to improve the generator, resulting in a convergence failure~\cite{goodfellow2014generative}. We intuitively hypothesize that if the discriminator is stronger at distinguishing samples in one group better than the others, it can provide more information in the gradient for the other group, leading this generator to converge more toward the group for which the discriminator provided more information. In the SVHN dataset, we created two groups by adding a black opacity layer over certain images (see section~\ref{sec:experiments}). We performed experiments to analyze the amount of signal the discriminator provides for each group in the training dataset. Figure \ref{fig:average_gradient_magni} showcases the average gradient norms of two groups present in the data during the training. We observe the disparities in the magnitude of their gradients is close to each other at the beginning of the training, and then enlarge over epochs. At the test time, we observed that the generator samples imbalanced distribution in which the group with higher gradient magnitude (shaded images) was over-represented (60\% vs 40\%). Note that in training data both groups were equally represented. Experiment with groups imbalance is presented in the section~\ref{sec:experiments}.
\subsubsection{Algorithm} Based on the previous empirical analysis, we can see the influence of the gradient norm of groups over the biased sampling at the test time. We designed a straightforward training process of GANs that controls the gradient of each group during training, such that the disparities in the magnitude of the gradient will not enlarges during the training. The key idea of our method is to modify the training of the process of the discriminator, by alternating groups used to optimize the discrimination across batches, and clipping the norm of their gradients when it exceeds a defined threshold (maximum gradient norm).  

\begin{algorithm}[tb]
\footnotesize
\caption{Training of the discriminator in RepFair-GAN (outline)}
\label{alg:cap}
\begin{algorithmic}
\Require{$t$ the batch size, $M$ training epochs, $C$ maximum gradient norm, and $\mathcal{L}_D(\Theta)$ the discriminator's loss function with network parameter $\Theta$.}
\For{$n \gets 1, M$}
    \State $U \gets True$
    \Comment{Boolean indicating whether to update with samples from group 0 or group 1}
    \For{each batch of size $t$}
    \If{U}
        \State $X_t\gets\{D | S=0\}$ 
        \Comment{random mini-batch of real samples from group 0}
    \Else
        \State $X_t\gets\{D | S=1\}$ 
        \Comment{random mini-batch of real samples from group 1} 
    \EndIf
    \State $g_t \gets \nabla_{\Theta}\mathcal{L}_D(\Theta; X_t)$
    \Comment{Compute the gradient using the discriminator's loss}
         
    \State $\bar{g_t} \gets \operatorname{clip}(g_t, C) $ 
    \Comment{Clip the gradient norm with max norm $C$}     
    
    \State $\Theta \gets \Theta - \alpha \bar{g_t}$ \Comment{Update network param.}
    \State $U \gets \operatorname{not} U$
    \State Generate samples from the generator $\bar{X_t}$ feed to the discriminator and take a gradient step.
    \EndFor
\EndFor
\end{algorithmic}
\end{algorithm}

Algorithm \ref{alg:cap} summarizes the training of the discriminator in our approach. We note that gradient clipping is an ingredient used in stochastic gradient descent (SGD) to solve different purposes. For instance, the gradient clipping is used to mitigate the gradient exploding problem, which arises when the weights of the model become very large so that it creates an overflow or underflow problem~\cite{lipton2015critical}. The clipping can be done on weights of models (value clipping) or on the norm of the gradient (norm clipping). Abadi et al.~\cite{abadi2016deep} used norm clipping to bound the influence of each training example to provide a privacy guarantee in SGD. In the same spirit, we apply of the norm clipping in our method to bound the gradient norm of different groups in the data such that none of these will provide an increasingly larger signal throughout the training of GAN. In our method, there is no a priori bound on
the size of the gradients, thus, the maximum gradient norm parameter ($C$) must be carefully chosen. A too small value of $C$ will avoid the generator to learn, due to the vanishing gradient problem, while a too-large value of $C$ will not change anything to training leading to the classic GAN's training with bias data generation.
In the next section, we show the effectiveness of our method on two datasets.  

\section{Experiments} 
\label{sec:experiments} 
To demonstrate the effectiveness of our approach, we experiment on two datasets in which we created two different groups with different representation rates. We trained GAN models using the classical training procedure and compared them to our training process (algorithm \ref{alg:cap}) in terms of the distribution of the generated groups at the time of test time over several training runs.
\subsection{Experiment settings}
\begin{figure}[tb]
\begin{multicols}{2} 
    \includegraphics[width=\linewidth]{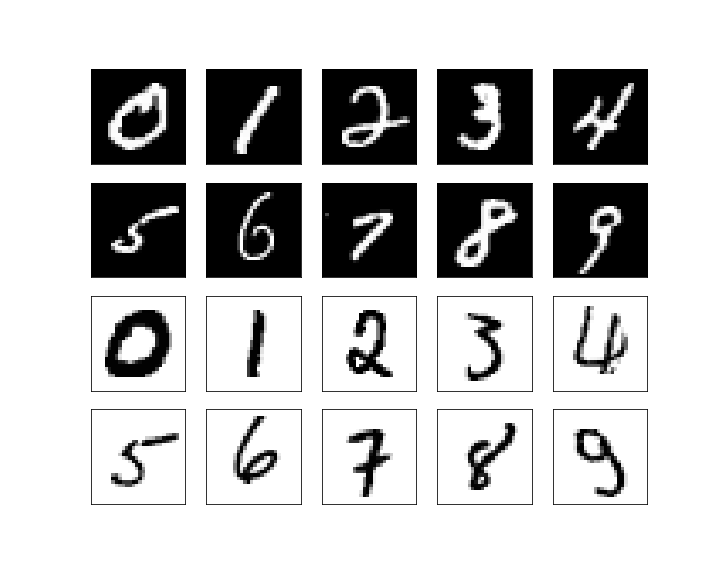}\par 
    \includegraphics[width=6cm]{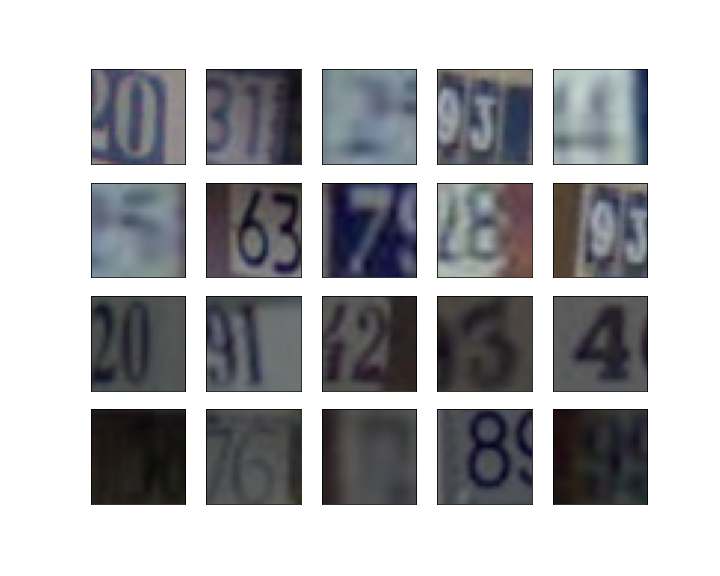}\par 
    \end{multicols} 
\caption{Example of two groups created from MNIST (left) and SVHN (right) datasets. In MNIST, one group of digits has a black background (two first rows) and another group has a white background (two last rows). In SVHN, one with normal images (first two rows), and for the second group we added a black overlay (shaded)}
\label{fig:datasets}
\end{figure}
\subsubsection{Datasets} 
We used MNIST~\cite{lecun1998gradient} and SVHN~\cite{netzer2011reading} datasets to run the experiments. MNIST dataset consists of 60,000 samples of labelled handwritten digits, for which we created two groups, one with white background color and the other with a black background color. SVHN contains 73,256 samples of home street numbers, which adopted a similar process to create groups, with the difference that the labels were not taken into consideration. As in~\cite{kenfack2021fairness}, the dataset was divided into two groups: the \textit{shaded} and the \textit{normal} SVHN images. For the shaded images, we added a shade filter on the photos and the other group contains the normal SVHN. Figure~\ref{fig:datasets} showcases examples of samples from each group for both datasets. We performed the experiments with balanced and imbalanced group representation.

\subsubsection{Models} For GAN models, we considered both the Vanilla and the CGAN. We trained Vanilla GAN on the SVHN dataset with different group representations and reported group distribution after sampling 10,000 points from the trained generator. We trained CGAN conditioned with class label on MNIST dataset and reported the results class-wise, i.e., group distribution for every type of digit. The objective was to examine whether the imbalance in group distribution affects all classes of digits or whether it is just an overall imbalance. All the models were trained with the classic training process and our proposed method. All generators were trained seven times, and the group distribution is averaged over all training runs.

For the model's architecture, for MNIST, we built a discriminator model whose input is a grey-scale $28\times28$ image and an integer representing the class label. The discriminator output is a binary prediction deciding if the generated image is real or fake. For our purpose, we implemented the discriminator neural network that follows GAN design common best practices such as choosing LeakyRelU as the activation function with slope 0.2 and $2\times2$ stride to downsample ~\cite{arjovsky2017towards}. For CGAN, the discriminator takes as input the image and the class label. The label will pass through a layer of embedding of size 50. Then the embedded layer is connected to the dense layers then is concatenated with the image before it goes through two convolutional layers with the same padding and stride of $2\times2$, each convolutional layer is followed by LeakyRelU with a slope 0.2. We used the Adam optimizer for SGD optimization with a learning rate of 0.0002 and a momentum of 0.5. 

The generator model handles the conditional architecture as well by having in the inputs the class label along with the noise vector sampled from Gaussian distribution. The label is handled as in the discriminator by passing it through the embedding layer of size 50, followed by a dense layer then concatenated with the noise vector before going through two transposed convolutional layers with the same padding and a stride of $2\times2$, followed by LeakyRelu with slope 0.2. the output of the generator model is a single $28\times28$ image generated by a convolutional layer with activation function $tanh$

Finally, the GAN model was simply built by sequentially providing the generator's output to the discriminator. The CGAN model also uses an Adam optimizer with a learning rate of 0.002 and a momentum of 0.5 and binary cross-entropy as a loss function. 

\subsubsection{Metrics} To evaluate the group distributions of the generated data, we trained classifiers on the real training data to predict the sensitive attribute. For the MNIST dataset, the classifier is trained to predict the background color of the digits, while for SVHN, it predicts whether the image is shaded or not. The classifier was trained on 70\% of the real images and achieved 100\% and {98}\% accuracy respectively for MNIST and SVHN on the testing dataset in SVHN. 
For MNIST the group was defined by checking the pixel intensity of the corner of the digit, i.e., if the intensity is greater than 0.5 we consider the digit as a black background and vice versa. We note that the classifier used to obtain the group distributions may also be biased, but for simplicity, we assume that this is not the case here. Furthermore, to quantify the unfairness, one can consider measuring the representation bias using the KL-divergence between the distribution of the generated samples and the uniform distribution (definition~\ref{def:representational-fairness}). In the results, we reported the frequency distribution of each group. Note that all experiments are based on 10 training runs, and frequencies are averaged across runs.

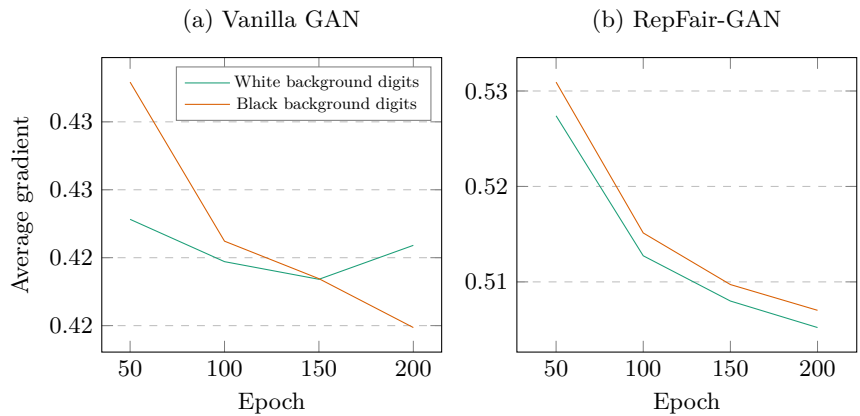
\begin{figure}[tb]
    \centering
    \begin{tikzpicture}
        \begin{groupplot}[
        group style={
                      group name=myplot,
                      group size= 2 by 1},
                      ]
        \nextgroupplot[
      legend style={ 
        draw=gray, 
        nodes={scale=0.7, transform shape}
        },   
    legend pos=north east,
    width=0.5\linewidth,
    height=5.5cm,
    xlabel={Epoch}, 
    ylabel={Average gradient},
    legend entries={White background digits,Black background digits},
    title={(a) Vanilla GAN}
    ]
    \addplot table [x=epoch, y=fair_white, col sep=comma]{images/csv/Average_gradient_for_white_background_digits_and_black_background_digits_Vanilla-GAN.csv};
    \addplot table [x=epoch, y=fair_black, col sep=comma]{images/csv/Average_gradient_for_white_background_digits_and_black_background_digits_Vanilla-GAN.csv};
    
    \nextgroupplot[
      legend style={ 
        draw=gray,  
        },  
    legend pos=south east,
    width=0.5\linewidth,
    height=5.5cm,
    xlabel={Epoch}, 
    enlarge x limits=.15, 
    title={(b) RepFair-GAN}
    ]
    \addplot table [x=epoch, y=fair_white, col sep=comma]{images/csv/Average_gradient_for_white_background_digits_and_black_background_digits_RepFair-GAN.csv};
    \addplot table [x=epoch, y=fair_black, col sep=comma]{images/csv/Average_gradient_for_white_background_digits_and_black_background_digits_RepFair-GAN.csv};
        
        \end{groupplot}
    \end{tikzpicture}
    \caption{Effect of the gradient clipping over the disparity in gradient norms of each group. Average gradient for white background digits and black background digits vanilla GAN. Clipping enforces the gradient norm of each group to remain closer during the training of the discriminator (b).}
    \label{fig:effect_clipping_mnist}
\end{figure} 

\subsection{Results} 
 
\subsubsection{Effect of gradient clipping}
During the training of CGAN, with reported the average gradient norm of each group across epochs. Figure~\ref{fig:effect_clipping_mnist} shows that in the unfair CGAN, the difference between the gradient magnitude of digits with white and black backgrounds tends to widen over epoch. In contrast, we observe that when using our fairness mechanism, the magnitude of the gradient of each group remains closer over epochs. We observed the same pattern on SVHN.

\subsubsection{Frequency distributions} 
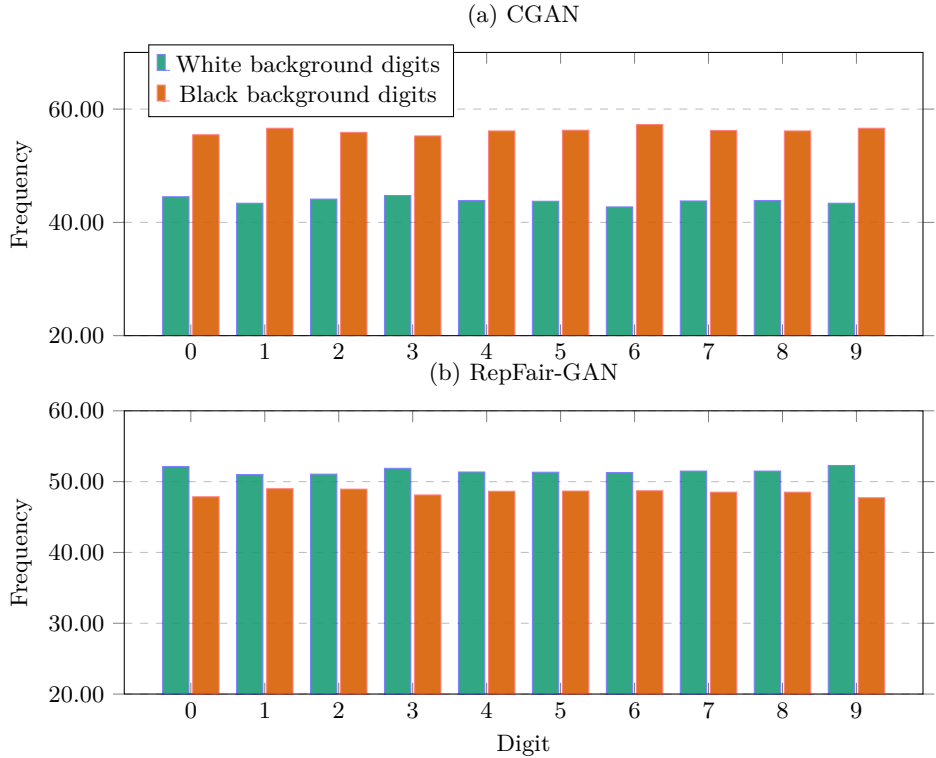
\begin{figure}[!ht]
    \centering
    \begin{tikzpicture}
        \begin{groupplot}[
        group style={
                      group name=myplot,
                      group size= 1 by 2},
                      ]
        \nextgroupplot[ 
        width=\textwidth,
        height=5.35cm,
        xtick={0,1,2,...,9},
         yticklabel style={%
            /pgf/number format/fixed,
            /pgf/number format/fixed zerofill,
            /pgf/number format/precision=2,
          },
          /pgfplots/error/.style={
            error bars/.cd,
            y dir=both, y explicit relative,
          },
          xbar=.05cm,
          enlarge x limits=.1,   
          /pgf/bar width=0.35cm,
            ymin=20,
            ymax=70,
            ylabel={Frequency},
            title={(a) CGAN},
            legend style={at={(0.03, .9)},anchor=west},
            legend entries={White background digits,Black background digits},
        ]
            \addplot+ [ybar=.5cm, fill=Dark2-A, fill opacity=.9, draw opacity=0.5] table [x index=0, y=White_backgorund, col sep=comma]{images/csv/Disterbution_normal-GAN.csv}; 
            \addplot+ [ybar=.5cm, fill=Dark2-B, fill opacity=.9, draw opacity=0.5] table [x index=0, y=Balck_backgorund, col sep=comma]{images/csv/Disterbution_normal-GAN.csv}; 
    
    \nextgroupplot[ 
        width=\textwidth,
        height=5.35cm,
        xtick={0,1,2,...,9},
         yticklabel style={%
            /pgf/number format/fixed,
            /pgf/number format/fixed zerofill,
            /pgf/number format/precision=2,
          },
          /pgfplots/error/.style={
            error bars/.cd,
            y dir=both, y explicit relative,
          },
          xbar=.05cm,
          enlarge x limits=.1,   
          /pgf/bar width=0.35cm,
            ymin=20,
            ymax=60,
            ylabel={Frequency},
            xlabel={Digit},
            title={(b) RepFair-GAN},
            ]
            \addplot+ [ybar=.5cm, fill=Dark2-A, fill opacity=.9, draw opacity=0.5] table [x index=0, y=White_backgorund, col sep=comma]{images/csv/Disterbution_RepFair-GAN.csv}; 
            \addplot+ [ybar=.5cm, fill=Dark2-B, fill opacity=.9, draw opacity=0.5] table [x index=0, y=Balck_backgorund, col sep=comma]{images/csv/Disterbution_RepFair-GAN.csv}; 
        \end{groupplot}
    \end{tikzpicture}
    \caption{Class-wise background color distribution of digits generated with CGAN trained on MNIST dataset. (a) Imbalanced digit color distributions for each class produced by CGAN. Digit colors were balanced in the training data and the generator favors generating digits with a black background at the test time. (b) Balanced digit color distributions for each class produced by RepFair-GAN. Our fairness mechanism provides a more uniform group representation at the test time.}
    \label{fig:digit_color_dist}
\end{figure}

We performed the experiments to analyze the representation bias in CGAN on MNIST dataset conditioned on the digits label. The model was trained for 200 epochs and the frequency of generating white background digits and black background digits the results are reported in figure~\ref{fig:digit_color_dist}(a) shows that even when groups are equally represented in the training set, the generator does not uniformly sample them at the test time. We perform the same experiment using the RepFair-GAN, the figure~\ref{fig:digit_color_dist}(b) shows the improvement in the uniform generation of all groups.

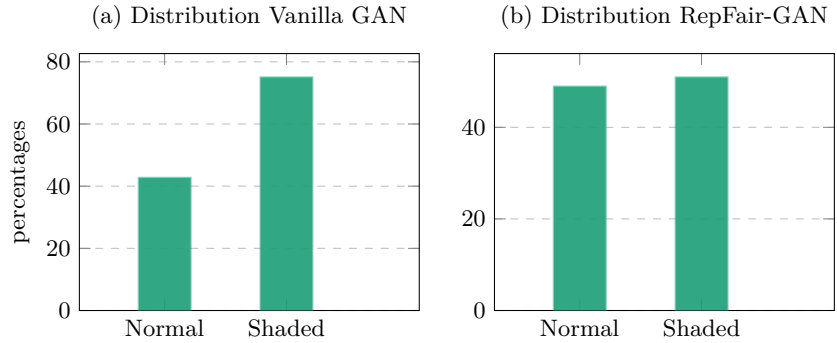
\begin{figure}[!ht]
    \centering
    \begin{tikzpicture}
        \begin{groupplot}[
        group style={
                      group name=myplot,
                      group size= 2 by 1}, 
                      ]
        \nextgroupplot[ 
        width=0.5\linewidth,
        height=5cm,
        xtick={0,1},
        xticklabels={Normal, Shaded},
        bar width = 20pt, 
        ymin=0,  
        xmax=2.09,
        xmin=-.7,
        ylabel={percentages},
        title={(a) Distribution Vanilla GAN}
        ]
            \addplot+ [ybar=.5cm, fill=Dark2-A, fill opacity=.9, draw opacity=0.5] table [x index=0, y=percentages, col sep=comma]{images/csv/Frequancy_of_each_image_type_normal_GAN.csv};
    
    \nextgroupplot[ 
        width=0.5\linewidth,
        height=5cm,
        xtick={0,1},
        xticklabels={Normal, Shaded},
        bar width = 20pt, 
        ymin=0,  
        xmax=2.09,
        xmin=-.7, 
        title={(b) Distribution RepFair-GAN}
        ]
            \addplot+ [ybar=.5cm, fill=Dark2-A, fill opacity=.9, draw opacity=0.5] table [x index=0, y=percentages, col sep=comma]{images/csv/Frequancy_of_each_image_type_FairRep-GAN.csv};
        \end{groupplot}
    \end{tikzpicture}
    \caption{Group distributions for vanilla GAN (a) and our method (b) on SVHN dataset trained with balanced group representation. Our method provides uniform generation compared to the vanilla GAN.}
    \label{fig:quantitave-svhn}
\end{figure}

\begin{figure}
\begin{multicols}{2} 
    \includegraphics[width=\linewidth]{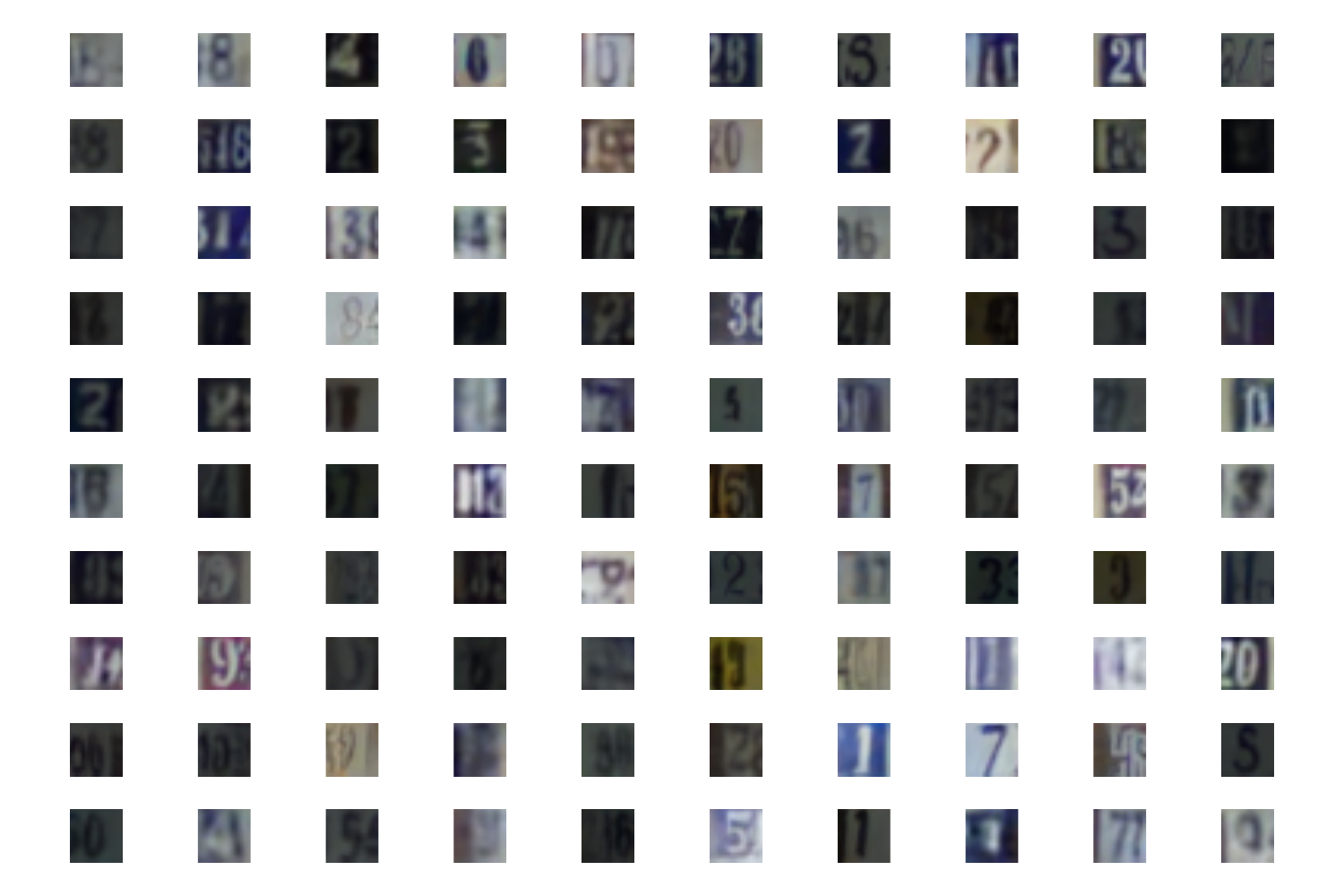}\par 
    \includegraphics[width=\linewidth]{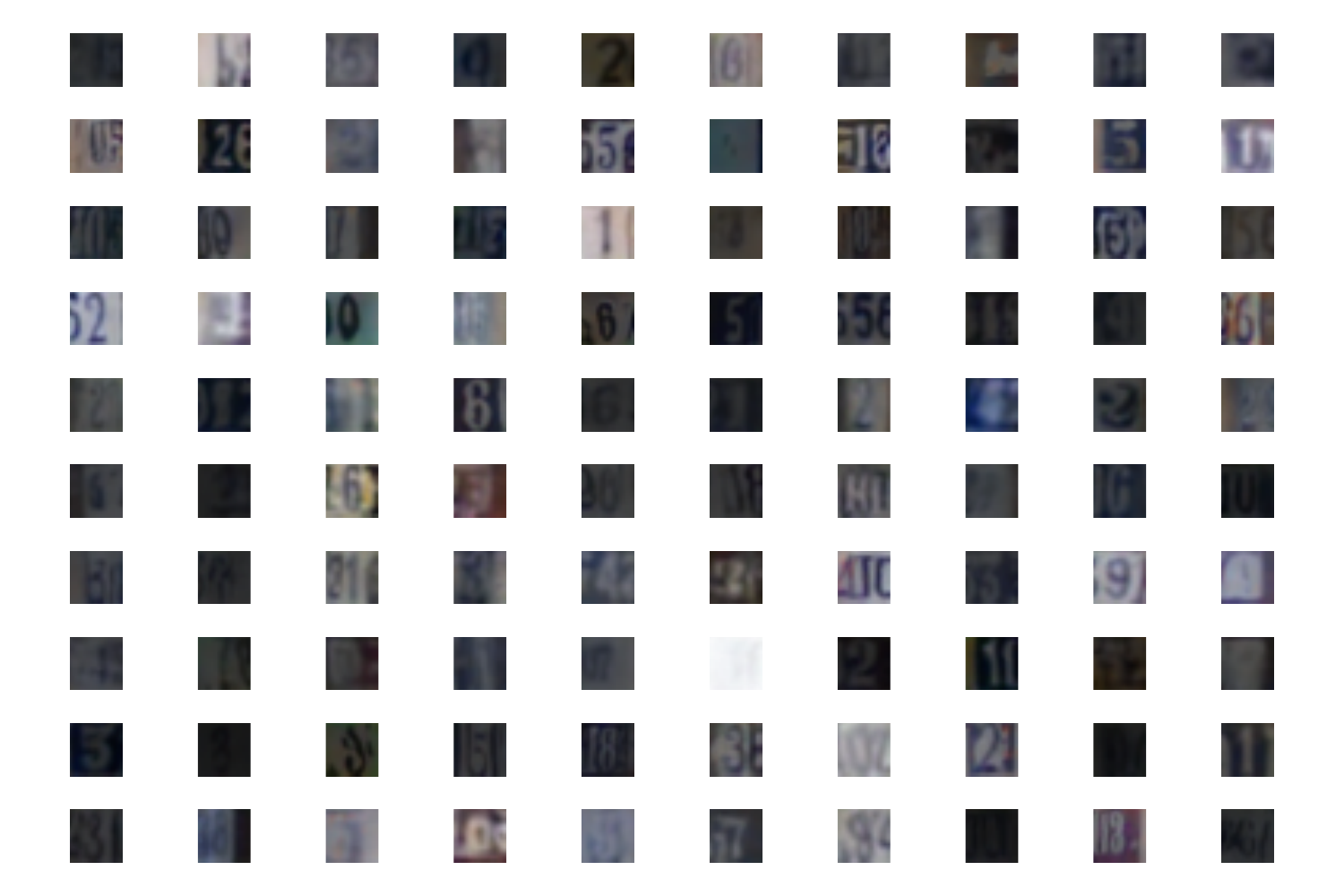}\par 
    \end{multicols} 
\caption{Qualitative results on SVHN dataset. Examples of images generated by GAN (left) and RepFair-GAN (right). We observe uniform generation of our method along with similar images quality}
\label{fig:qualitative-svhn}
\end{figure}

\begin{figure}
\begin{multicols}{2} 
    \includegraphics[width=\linewidth]{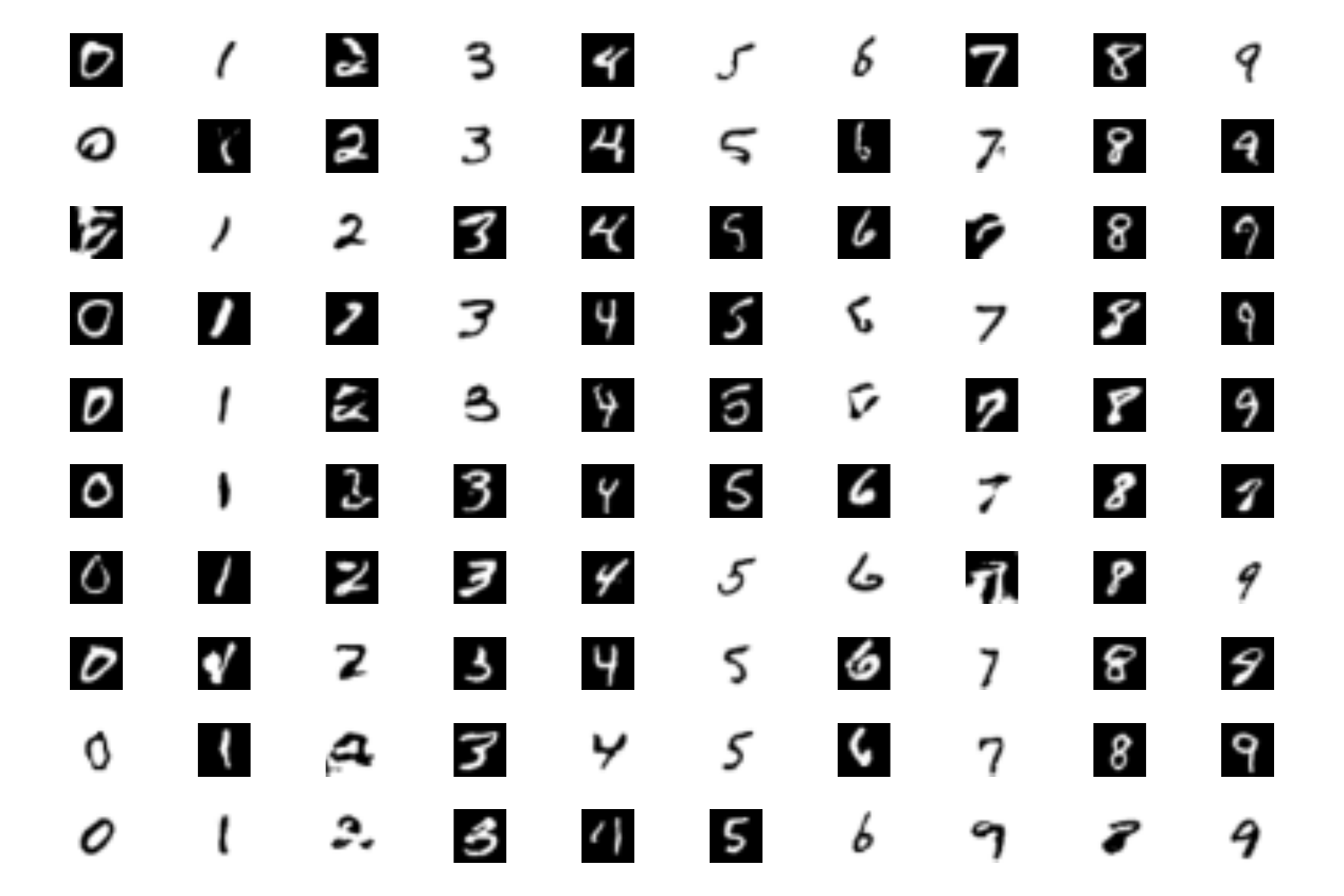}\par 
    \includegraphics[width=\linewidth]{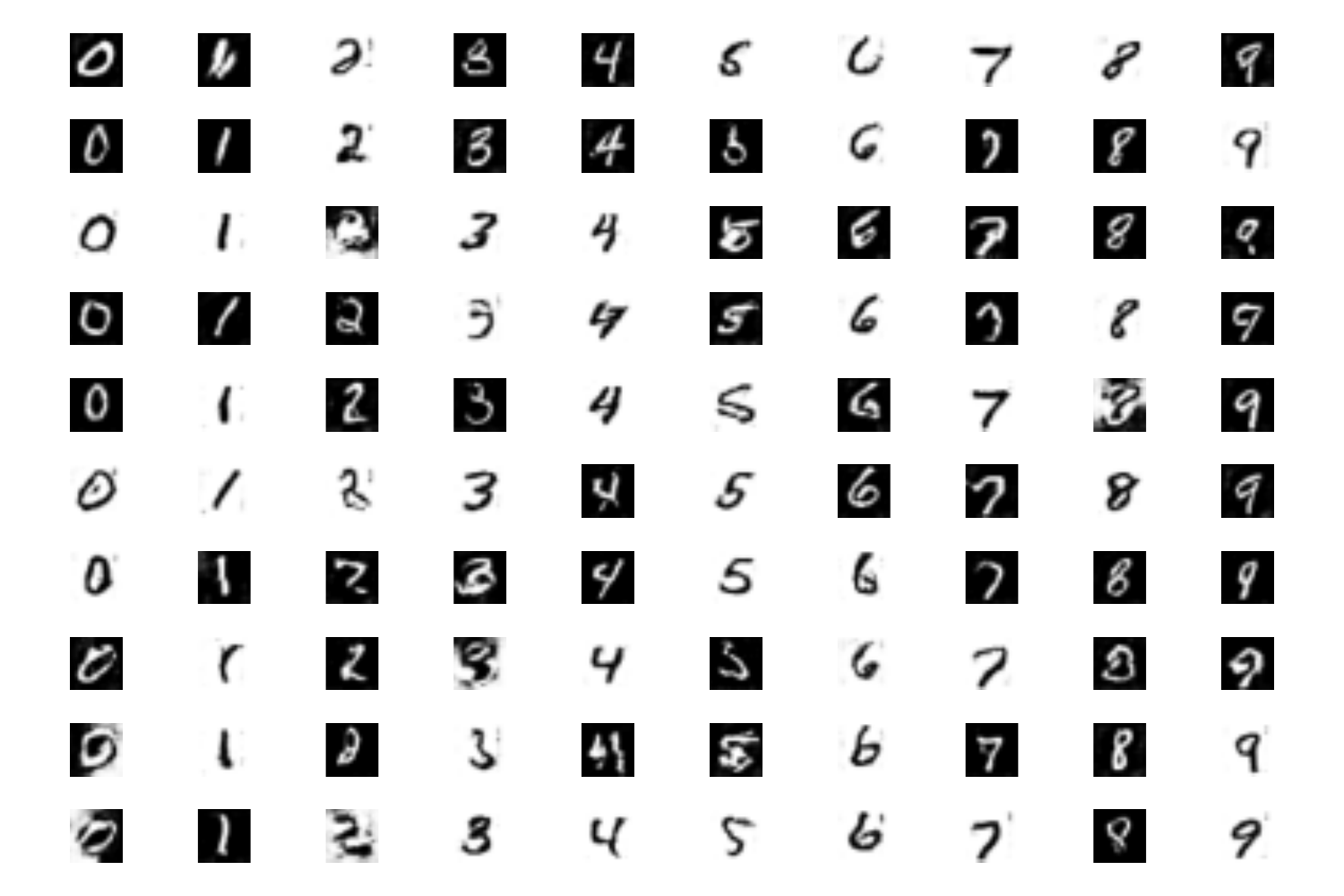}\par 
    \end{multicols} 
\caption{Qualitative results on MNIST dataset. Examples of digits generated by CGAN (left) and RepFair-GAN (right). We observe that for each type of digit, group are equally represented with our method while preserving the same quality as  the classic GAN}
\label{fig:qualitative-cgan}
\end{figure}

We also analysed if the clipping can affect the quality of the generated images. Figure~\ref{fig:qualitative-cgan} shows that our method can uniformly generate all groups while preserving image quality. For each class, we observed that all colors were equally represented while on CGAN without our fairness mechanism, digits with black backgrounds over-represented. Our method reduces the difference in group representation from about 18\% to 3\%.    Similar results were observed on SVHN datasets Cf. figures~\ref{fig:quantitave-svhn} and~\ref{fig:qualitative-svhn}. This shows that by controlling the magnitude of the gradient of each group, the discriminator is no more competent at distinguishing real samples from one group than from the other, and thus, pushes the generator to not converge primarily to the distribution of one group while neglecting the other.

\subsubsection{Effect of maximum gradient norm}
\begin{figure}[!ht]
    \centering
    \begin{tikzpicture}
        \begin{axis}[ 
        width=\textwidth,
        height=5cm,
        xtick={0,1,...,9},
        xticklabels={0,0.5,1,2,10,20,50,100,200},
         yticklabel style={%
            /pgf/number format/fixed,
            /pgf/number format/fixed zerofill,
            /pgf/number format/precision=2,
          },
          /pgfplots/error/.style={
            error bars/.cd,
            y dir=both, y explicit relative,
          },
          xbar=.05cm,
          enlarge x limits=.1,   
          /pgf/bar width=0.35cm,
            ymin=20,
            ymax=86,
            ylabel={Frequency},
            title={Digit background distributions},
            legend entries={White background digits,Black background digits},
            xlabel={Maximum gradient norm ($C$)}
            ]
            \addplot+ [ybar=.5cm, fill=Dark2-A, fill opacity=.9, draw opacity=0.5] table [x index=0, y=White_backgorund, col sep=comma]{images/csv/RepFair-GAN_representaion_with_different_norms.csv}; 
            \addplot+ [ybar=.5cm, fill=Dark2-B, fill opacity=.9, draw opacity=0.5] table [x index=0, y=Balck_backgorund, col sep=comma]{images/csv/RepFair-GAN_representaion_with_different_norms.csv}; 
        \end{axis} 
    \end{tikzpicture}
    \caption{Group distributions with different maximum gradient norms in RepFair-GAN on MNIST data. For max norm less than 2 the training does not converge due to the vanishing gradient.  For max gradient norms between 2 and 10 groups are sampled uniformly and the sampling becomes biased for values bigger than 20.}
    \label{fig:representaion_with_different_norms}
\end{figure}
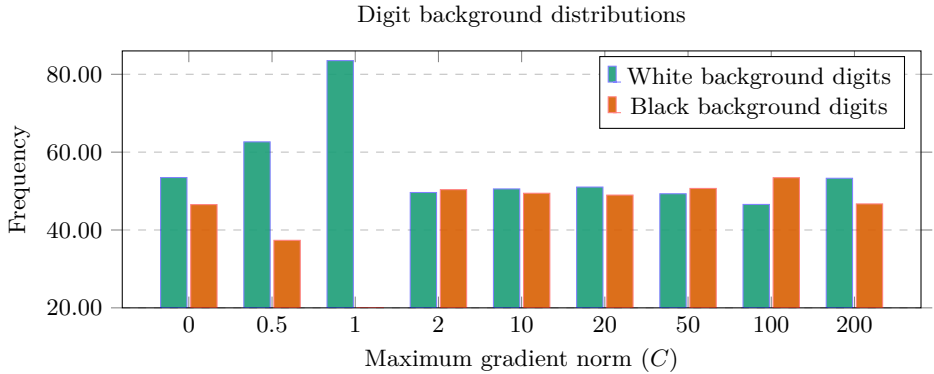

We trained CGAN on MNIST with our fairness mechanism using different values of maximum gradient norms, $C \in \{0.1, 0.5, 1, 2, 10, 50, 100\} $. We reported the overall group distribution, i.e., for all classes. In figure~\ref{fig:representaion_with_different_norms}, we observed that for $C = $ 0.1, 0.5, and 1 the clipping was too high and the model was unable to converge properly, and the generator could not produce realistic digits leading to almost random results in the distribution, i.e., images were a mixture of white and black backgrounds. For norm $C$ between 2 and 10 the clipping was good enough to let the model converge and to provide fair sampling at the test time. For bigger values of $C$ ($\geq 20$ ), we observed that the representation bias started occurring, suggesting that the maximum gradient value was too big and did not influence the training. In all other experiments, on MNIST and SVHN with $C=2$. The best value $C$ can be chosen via hyper-parameter by turning on a validation set.

\subsubsection{Effect of group imbalance}
\begin{figure}[tb]
    \centering
    \begin{tikzpicture}
        \begin{groupplot}[
        group style={
                      group name=myplot,
                      group size= 2 by 3},
                      ]
                      
        \nextgroupplot[ 
        width=0.5\linewidth,
        height=4.5cm,
        xtick={0,1},
        xticklabels={Normal, Shaded},
        bar width = 20pt, 
        ymin=0,  
        xmax=2.09,
        xmin=-.7,
        ylabel={percentages},
        title={(a) 20\% vs 80\% (GAN)}
        ]
            \addplot+ [ybar=.5cm, fill=Dark2-A, fill opacity=.9, draw opacity=0.5] table [x index=0, y=percentages, col sep=comma]{images/csv/20_resutls_vanilla.csv};
        
        \nextgroupplot[ 
        width=0.5\linewidth,
        height=4.5cm,
        xtick={0,1},
        xticklabels={Normal, Shaded},
        bar width = 20pt, 
        ymin=0,  
        xmax=2.09,
        xmin=-.7, 
        title={(b) 20\% vs 80\% (RepFair-GAN)}
        ]
            \addplot+ [ybar=.5cm, fill=Dark2-A, fill opacity=.9, draw opacity=0.5] table [x index=0, y=percentages, col sep=comma]{images/csv/20_resutls_RepFair.csv};
   \vspace{1.2cm}%
     \nextgroupplot[ 
        width=0.5\linewidth,
        height=4.5cm,
        xtick={0,1},
        xticklabels={Normal, Shaded},
        bar width = 20pt, 
        ymin=0,  
        xmax=2.09,
        xmin=-.7,
        ylabel={percentages},
        title={30\% vs 70\%}
        ]
            \addplot+ [ybar=.5cm, fill=Dark2-A, fill opacity=.9, draw opacity=0.5] table [x index=0, y=percentages, col sep=comma]{images/csv/30_resutls_vanilla.csv};
        
        \nextgroupplot[ 
        width=0.5\linewidth,
        height=4.5cm,
        xtick={0,1},
        xticklabels={Normal, Shaded},
        bar width = 20pt, 
        ymin=0,  
        xmax=2.09,
        xmin=-.7, 
        title={30\% vs 70\%}
        ]
            \addplot+ [ybar=.5cm, fill=Dark2-A, fill opacity=.9, draw opacity=0.5] table [x index=0, y=percentages, col sep=comma]{images/csv/30_resutls_RepFair.csv};
        
        \nextgroupplot[ 
        width=0.5\linewidth,
        height=4.5cm,
        xtick={0,1},
        xticklabels={Normal, Shaded},
        bar width = 20pt, 
        ymin=0,  
        xmax=2.09,
        xmin=-.7,
        ylabel={percentages},
        title={40\% vs 60\%}
        ]
            \addplot+ [ybar=.5cm, fill=Dark2-A, fill opacity=.9, draw opacity=0.5] table [x index=0, y=percentages, col sep=comma]{images/csv/40_resutls_vanilla.csv};
        
        \nextgroupplot[ 
        width=0.5\linewidth,
        height=4.5cm,
        xtick={0,1},
        xticklabels={Normal, Shaded},
        bar width = 20pt, 
        ymin=0,  
        xmax=2.09,
        xmin=-.7, 
        title={40\% vs 60\%}
        ]
            \addplot+ [ybar=.5cm, fill=Dark2-A, fill opacity=.9, draw opacity=0.5] table [x index=0, y=percentages, col sep=comma]{images/csv/40_resutls_RepFair.csv};
            
        \end{groupplot}
    \end{tikzpicture}
    \caption{Group distributions on SVHN with different group representation ratios 20\%, 30\%, and 40\% respectively on each row. Distribution on the left column is obtained from a classic vanilla GAN and distribution on the right are obtained from the Vanilla GAN trained with our fairness mechanism (RepFair-GAN).}
    \label{fig:quantitave-svhn-imbalance}
\end{figure}
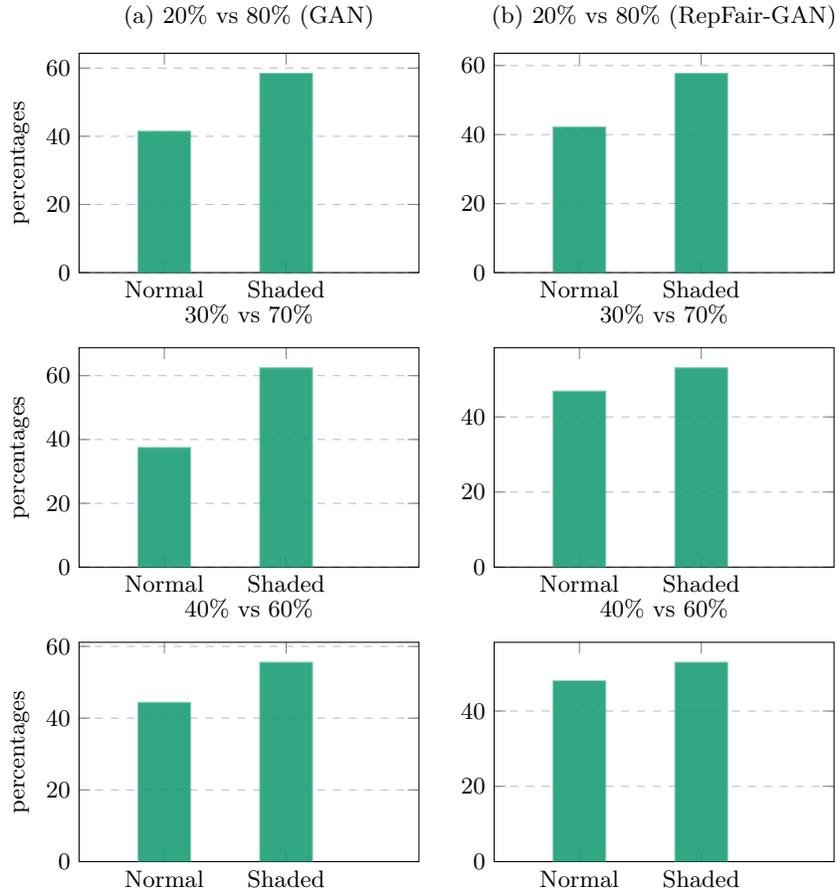

We trained vanilla GANs and RepFair-GAN on SVHN with different group imbalance rates: 20\%, 30\%, 40\%. As shown in the figure~\ref{fig:quantitave-svhn-imbalance}, vanilla GAN generates data highly biased towards the well-represented group (shaded image) in all cases. While distributions provided by our method are less biased. Surprisingly the distribution at 20\% is less skewed than that of 30\%. In all cases, a default value $C=2$ was used. Tuning this value on a validation dataset for each case would result in better fairness performance.    

 s
\section{Conclusion and future work}
\label{ref:conclusion}
In this work, we shed light on the reason for unfairness in GANs. Using empirical evidence,  we attributed this phenomenon to the disparity in groups' gradient norms during the discriminator training. We proposed a new training for GAN that mitigates the representation bias of the generator at the test time. In a nutshell, our method controls the gradient of each group during the training by alternating their gradient step within batches and clipping their norms if it exceeds a defined maximum value. Our experiments demonstrated the effectiveness of this process in generating groups present in data uniformly even when groups are imbalanced in the dataset. In future work, we will test our method against existing techniques for fair data generation while considering more complex and high-dimensional data. Furthermore, this work laid the foundation for bias analysis and mitigation in ML from the perspective of gradient disparity. It is therefore important to examine the applicability of the proposed fairness mechanism to other gradient-based models and classification tasks. Also, the relationship between fair data sampling and diversity in GANs remains a research direction that needs to be explored.   
%
%
%
\bibliographystyle{splncs04}
\bibliography{paper}
\end{document}